\documentclass[english, natbib]{acm_proc_article-sp}
\usepackage[T1]{fontenc}
\usepackage[latin9]{inputenc}
\usepackage{booktabs}
\usepackage{graphicx}
\usepackage{multirow}
\usepackage{amsmath}
\usepackage{amssymb}
\usepackage[numbers]{natbib}

\makeatletter

\pdfpageheight\paperheight
\pdfpagewidth\paperwidth

\providecommand{\tabularnewline}{\\}

\usepackage{babel}

\makeatother

\usepackage{babel}
\begin{document}
\newfont{\mycrnotice}{ptmr8t at 7pt}
\newfont{\myconfname}{ptmri8t at 7pt} 
\let\crnotice\mycrnotice% 
\let\confname\myconfname%

\crdata{XXXXX}

\clubpenalty=10000     
\widowpenalty = 10000

\numberofauthors{1}
\author{ \alignauthor Vin\'{i}cius Veloso de Melo\\ \affaddr{Institute of Science and Technology (ICT)}\\ \affaddr{Federal University of S\~{a}o Paulo (UNIFESP)}\\ \affaddr{S\~{a}o Jos\'{e} dos Campos, SP, Brazil}\\ \email{vinicius.melo@unifesp.br} }

\title{A novel metaheuristic method for solving constrained engineering
optimization problems: Drone Squadron Optimization}

\maketitle

\begin{abstract}
Several constrained optimization problems have been adequately solved
over the years thanks to advances in the metaheuristics area. In this
paper, we evaluate a novel self-adaptive and auto-constructive metaheuristic
called Drone Squadron Optimization (DSO) in solving constrained engineering
design problems. This paper evaluates DSO with death penalty on three
widely tested engineering design problems. Results show that the proposed
approach is competitive with some very popular metaheuristics.
\end{abstract}


\maketitle

\section{Introduction}

Several real-world engineering design problems can be formulated in
a nonlinear programming way, where one wants to find a solution $\vec{x}$
that optimizes $f(\vec{x})$ 

\begin{equation}
\textrm{subject\,\ to}\,\begin{array}{cc}
h_{i}(\vec{x})=0 & i=1,2,...,m\\
g_{i}(\vec{x})\leqslant0 & i=1,2,...,P
\end{array}
\end{equation}

\noindent where $f(\vec{x})$ is the objective function to be optimized,
and $\vec{x}\in\Re^{n}$ is an $n$-dimensional vector $\vec{x}=[x_{1},x_{2},...,x_{n}]^{T}$.
Each $x_{k},\,k=1,...,n$ can be bounded by lower and upper limits
$L_{i}\leqslant x_{k}\leqslant U_{i}$; $h_{i}(\vec{x})$ and $g_{i}(\vec{x})$
are the equality and inequality constraints, respectively; $m$ is
the number of equality constraints and $p$ is the number of inequality
constraints, where both can be linear or nonlinear. Constraints reduce
the feasible search-space of the problem and, instead of making it
easier, make it more difficult because feasible solutions can be hard
to find. Thus, algorithms able to solve this kind of task are welcome
in engineering and manufacturing processes.

Many metaheuristics have been proposed to solve constrained problems,
being a very active research topic. Most of the proposed algorithms
are nature-inspired~\citep{coello2000use,ray2003society,akay2012artificial,sadollah2013mine,askarzadeh2016novel},
some are hybrid approaches~\citep{Michalewicz96evolutionaryalgorithms,He:2007:ECP:1219187.1219657,He20071407,Liu:2010:HPS:1660170.1660493},
some are classical algorithms with new operators~\citep{DBLP:journals/eswa/Coelho10,He:2007:ECP:1219187.1219657,Pant:2009:LDI:1670662.1670669,DBLP:journals/aei/CoelloM02},
others are self-adaptive version of classical algorithms~\citep{Coello2000,Michalewicz96evolutionaryalgorithms,Mezura2006,4358768}.

In this work, we investigated Drone Squadron Optimization (DSO\footnote{https://github.com/melovv/DSO-MATLAB}~\citep{de2017drone}),
a recently proposed self-adaptive metaheuristic which is self-improved
online by a hyper-heuristic. DSO is an artifact-inspired technique,
as opposed to many algorithms used nowadays, which are nature-inspired.
DSO is very flexible because it is not related to behaviors or natural
phenomena. Therefore, it can mimic any behavior.

The paper is organized as follows: In Section~2, we briefly introduce
the DSO algorithm. Section~3 provides the description of the idea
to handle constraints. Section~4 presents the numerical examples
(engineering problems), details of the experiments, the results obtained
and the discussion. Finally, in Section~5 some conclusions are drawn
about the results.

\section{Drone Squadron Optimization}

Drones can navigate remotely o completely autonomously. They have
sensors, can communicate over large distances and, one of the most
important features: can be upgraded/ improved not only in terms of
hardware but also by changing their software~(\emph{the firmware}).
Therefore, as a software controls their behavior, researchers are
free to add any kind of mechanism to the algorithm as a regular software
upgrade. Thus, there is no need to look for a natural phenomena to
justify the improvement.

The Drone Squadron Optimization (DSO) is based on the movement of
entities on the search-space. However, as explained before, the movement
of the squadron is not necessarily based on behavior observed in nature.
DSO's approach allows it to automatically choose to use recombination
and/or perturbation of solutions with distinct procedures, making
it act as an evolutionary algorithm, swarm algorithm, probabilistic
algorithm, or other, according to how it performs on the search landscape.
Moreover, those procedures may have their \emph{actual code} updated
during the search.

DSO has two core parts: the semi-autonomous drones that fly over a
landscape to explore, and the Command Center that processes the retrieved
data and updates the drones' firmware whenever necessary. The self-adaptive
aspect of DSO in this work is the perturbation/movement scheme, which
is the function used to generate target coordinates (solutions). This
function is evolved by the Command Center during the global optimization
process in order to automatically adapt DSO to the search landscape,
trying to increase the search efficacy.

The DSO algorithm presented here is composed of one Drone Squadron
with different teams and a Command Center, which uses information
collected from the drones to maintain partial control of the search,
and to develop new firmware to control the drones. A drone \emph{is
not} a solution; it moves to a \emph{coordinate} which is a solution.
A drone has a firmware containing the functions (codes) and configurations
used by the teams to search the landscape. All drones in the same
team share a firmware, but they can be located in different regions
of the search-space. In this work, the perturbation function is an
actual source code; it is a \emph{string} to be parsed and executed
by the drone. 

In~\citep{de2017drone}, DSO was proposed to solve unconstrained
(box-con-strained) numerical optimization problems. To solve constrained
optimization problems, DSO must employ a constraint handling mechanism.

\section{Constraints handling}

Constraints handling is an important issue in constrained optimization.
Such mechanism must guide the optimization technique into feasible
regions and be able to reach the bounds of the search-space. A general,
but usually not recommended, approach when metaheuristics are used
to solve constrained problems is the adoption of penalties~\citep{michalewicz1995survey}.
A penalty function (see Equation~\ref{eq:penalty}) can be applied
to unfeasible solutions to generate a poor function value. If the
solution~(\emph{$\vec{x}$}) is feasible~($F$), then the penalty
is not applied. In minimization problems, we add a penalty. Otherwise,
in maximization problems, we subtract a penalty.

\begin{equation}
f(\vec{x})=\begin{cases}
\begin{array}{c}
objfun(\vec{x})\\
objfun(\vec{x})+penalty(\vec{x})
\end{array} & \begin{array}{c}
if\,\vec{x}\,\in\,F\\
otherwise
\end{array}\end{cases}\label{eq:penalty}
\end{equation}

Using this approach, an unfeasible solution can be dropped from the
population in the next iteration of the algorithm, justifying the
common name of Death Penalty~\citep{michalewicz1995survey}. This
allows the constrained problem to be treated as an unconstrained one.
However, it does not allow to differentiate two unfeasible solutions
as both get the same $f(\vec{x})$. While this characteristic turns
unfeasible regions into plateaus, it is the simplest constraint handling
mechanism and may be useful in some problems.

\section{Experimental analysis}

In this paper, we investigate three well-explored engineering design
problems: the design of a Welded Beam, a Speed Reducer, and a Three-bar
truss. The definitions of these problems can be seen in related work~\citep{askarzadeh2016novel}.

\subsection{Computational Environment}

DSO was implemented in Matlab (R) 2010, compatible with Octave. The
experiments were run on an Intel (R) i7 6700k, Arch Linux 4.11.9-1-ARCH.

\subsection{Configuration}

As the problems investigated in this work are minimization ones, the
penalty function simply returns the maximum value accepted by Matlab:
$realmax=1.79769313486232e+308$. Thus, all unfeasible solutions have
$f(\vec{x})=realmax$.

The maximum number of evaluations is: 30,000 for the Welded Beam problem,
30,000 for the Pressure Vessel problem, and 3000 for the Three-bar
truss problem. We performed 50 independent runs. All test problems
were solved using the following set of parameters~\citep{de2017drone}:
$Teams=4$,$\break$ $Drones\_per\_team=15$, $C1=0.5$, $C2=0.3$,
$C3=0.7$, $MaxStagnation=50$, $Pacc=0.5$, $Commander\_iter=2$,
$ConvThres=1e\textrm{{-}}8$.

\subsection{Results}

Tables \ref{tab:Results-DSO-Welded}, \ref{tab:Results-DSO-Pressure},
and \ref{tab:Results-DSO-Bar} have statistics comparing DSO and related
methods on the investigated problems. As one may notice, DSO found
the same as or better solutions than the other methods. The average
solution was not as good, probably because DSO performed much fewer
evaluations and had outliers or because of the simple death penalty
approach. Nevertheless, it is important to remember that DSO evolves
the firmware, but one cannot guarantee that the new functions are
useful. Therefore, poor-quality or invalid functions may be generated. 

\begin{table}[!h]
\caption{\label{tab:Results-DSO-Welded}Statistics of best results found by
DSO for the Welded beam problem.}

\resizebox{1\columnwidth}{!}{
\centering{}%
\begin{tabular}{cccc}
\toprule 
\multirow{1}{*}{Method} & Evaluations & Best & Average\tabularnewline
\midrule
DSO & 30,000 & 1.72485230859736 & 1.82878489196467\tabularnewline
ABC~\citep{akay2012artificial} & 30,000 & 1.724852 & 1.741913\tabularnewline
CSA~\citep{askarzadeh2016novel} & 100,000 & 1.7248523086 & 1.7248523086\tabularnewline
GA~\citep{coello2000use} & 900,000 & 1.748309 & 1.771973\tabularnewline
MBA~\citep{sadollah2013mine} & 47,370 & 1.724853 & 1.724853\tabularnewline
PSO-DE~\citep{liu2010hybridizing} & 66,600 & 1.724852 & 1.724852\tabularnewline
SC~\citep{ray2003society} & 33,095 & 2.3854347 & 3.0025883\tabularnewline
\bottomrule
\end{tabular}}
\end{table}

\begin{table}[!h]
\caption{\label{tab:Results-DSO-Pressure}Statistics of best results found
by DSO for the Pressure Vessel problem. Symbol '-' means Not Available.}

\resizebox{1\columnwidth}{!}{
\centering{}%
\begin{tabular}{cccc}
\toprule 
\multirow{1}{*}{Method} & Evaluations & Best & Average\tabularnewline
\midrule
DSO & 30,000 & 5885.3332019268 & 6489.2853259488\tabularnewline
ABC~\citep{akay2012artificial} & 30,000 & 6059.714736 & 6245.308144\tabularnewline
CSA~\citep{askarzadeh2016novel} & 250,000 & 6059.71436343 & 6342.49910551\tabularnewline
GA~\citep{coello2000use} & 900,000 & 6288.7445 & 6293.8432\tabularnewline
MBA~\citep{sadollah2013mine} & 70,650 & 5889.3216 & 6200.64765\tabularnewline
PSO-DE~\citep{liu2010hybridizing} & 42,100 & 6059.714 & 6059.714\tabularnewline
SC~\citep{ray2003society} & - & - & -\tabularnewline
\bottomrule
\end{tabular}}
\end{table}

\begin{table}[!h]
\caption{\label{tab:Results-DSO-Bar}Statistics of best results found by DSO
for the Three-bar truss problem.}

\resizebox{1\columnwidth}{!}{
\centering{}%
\begin{tabular}{cccc}
\toprule 
\multirow{1}{*}{Method} & Evaluations & Best & Average\tabularnewline
\midrule
DSO & 3000 & 263.895843376498 & 264.067092887924\tabularnewline
ABC~\citep{akay2012artificial} & - & - & -\tabularnewline
CSA~\citep{askarzadeh2016novel} & 25,000 & 263.8958433765 & 263.8958433765\tabularnewline
GA~\citep{coello2000use} & - & - & -\tabularnewline
MBA~\citep{sadollah2013mine} & 13,280 & 263.895852 & 263.897996\tabularnewline
PSO-DE~\citep{liu2010hybridizing} & 17,600 & 263.895843 & 263.895843\tabularnewline
SC~\citep{ray2003society} & 17,610 & 263.895846 & 263.903356\tabularnewline
\bottomrule
\end{tabular}}
\end{table}

\section{Conclusions}

In this paper, we evaluated Drone Squadron Optimization with a death
penalty function to solve three well-known constrained engineering
design problems. Experiments were conducted on the design of: a welded
beam, a pressure vessel, and a three-bar truss. Results show that
DSO was able to achieve the best known solution of each problem after
a relatively small number of function evaluations.

When compared to nature-inspired approaches, DSO with a simple penalty
function found equal or better solutions. We intend to improve the
technique using a better constraint handling mechanism to reduce the
average solution quality.

\section{Acknowledgments}

I'm still trying to get a grant...

\bibliographystyle{abbrv}

\end{document}